\documentclass[letterpaper, 10 pt, conference]{ieeeconf}

\IEEEoverridecommandlockouts                              
\usepackage{epsfig}           
\usepackage{mathrsfs}
\usepackage{enumerate}
\usepackage{latexsym}
\usepackage{eufrak}
\usepackage{multirow}
\usepackage{makecell}
\usepackage{array}
\usepackage{graphicx}
\usepackage{epstopdf}

\usepackage{setspace}
\usepackage[hidelinks]{hyperref}
\usepackage[linesnumbered,ruled,vlined]{algorithm2e}

\usepackage{color}
\usepackage{amsmath}
\usepackage{amsfonts}
\usepackage{amssymb}
\usepackage{bm}
\usepackage{wrapfig}

\newlength\mylen

\renewcommand \paragraph[1] {\vspace{0.05cm} \textbf{#1}}

\def\etal{\emph{et al.}\xspace}

\usepackage{color}

\title{\textbf{Ironing of Deformable Objects using Surface Analysis	}}
\title{\textbf{Multi-Sensor Surface Analysis for Robotic Ironing	}}

\small {
\author{Yinxiao Li, Xiuhan Hu, Danfei Xu, Yonghao Yue, Eitan Grinspun, Peter K. Allen
\thanks{All the authors are with Department Computer Science, Columbia University, New York, NY, USA \tt\small \{yli@cs., xh2234, dx2143, yonghao@cs.,  eitan@cs. allen@cs.\}columbia.edu}
}
}

\begin{document}
\maketitle
\thispagestyle{empty}
\pagestyle{empty}


\begin{abstract}

Robotic manipulation of deformable objects remains a challenging task.
One such task is to iron a piece of cloth autonomously. Given a roughly 
flattened cloth, the goal is to have an ironing plan that can iteratively apply a 
regular iron to remove all the major wrinkles by a robot. We present a novel 
solution to analyze the cloth surface by fusing 
two surface scan techniques: a curvature scan and a discontinuity scan.
The curvature scan can estimate the height deviation of the cloth surface, 
while the discontinuity scan can effectively detect sharp surface features, such as wrinkles.
We use this information to detect the regions that need to be pulled and extended before ironing, 
and the other regions where we want to detect wrinkles and apply ironing to remove the wrinkles. 
We demonstrate that our hybrid scan technique is able to capture and classify wrinkles over the surface robustly.
Given detected wrinkles, we enable a robot to iron them using shape features.
Experimental results show that using our wrinkle analysis algorithm,
our robot is able to iron the cloth surface and effectively remove the wrinkles.

\end{abstract}

\section{Introduction}
\label{sec:intro}

\begin{figure}[t]
\begin{center}
\includegraphics[width=0.48\textwidth]{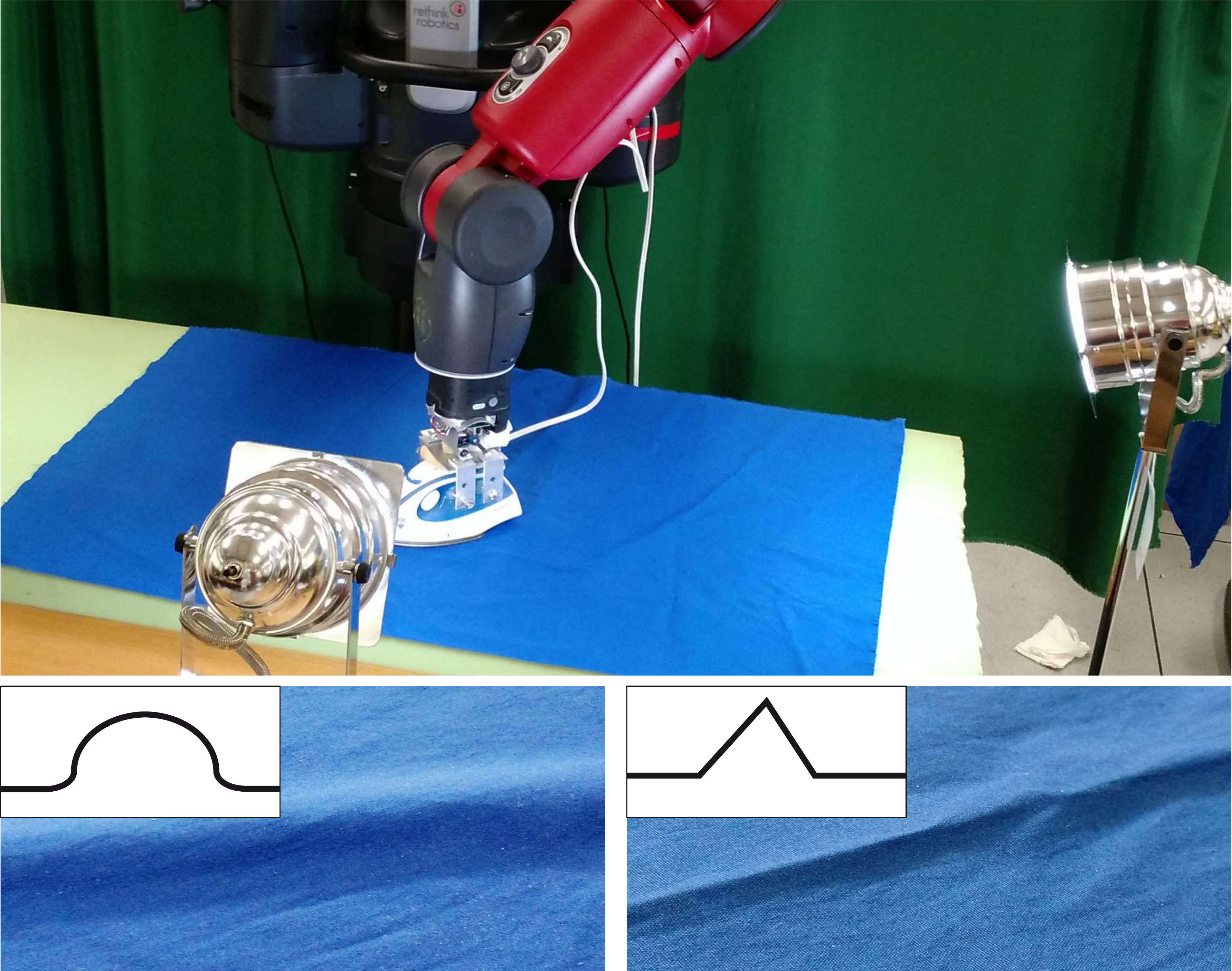}
\end{center}
   \caption{
	{\sc{top:}} A Baxter robot is ironing a piece of cloth on the table, assisted by a Kinect sensor and two light sources.
	{\sc{bottom:}} Samples of smooth region (left) and non-smooth region (right). The smooth regions are defined as height bumps, which can be removed by iterative pulling the cloth boundary. 
	The non-smooth regions are defined as wrinkles which can be removed by ironing. 
There is also a Kinect sensor on top of the table to capture depth and color images (not shown).}
\label{fig:iron_intro}
\end{figure}

\begin{figure*}[t]
  \centering
  \includegraphics[width=0.99\textwidth]{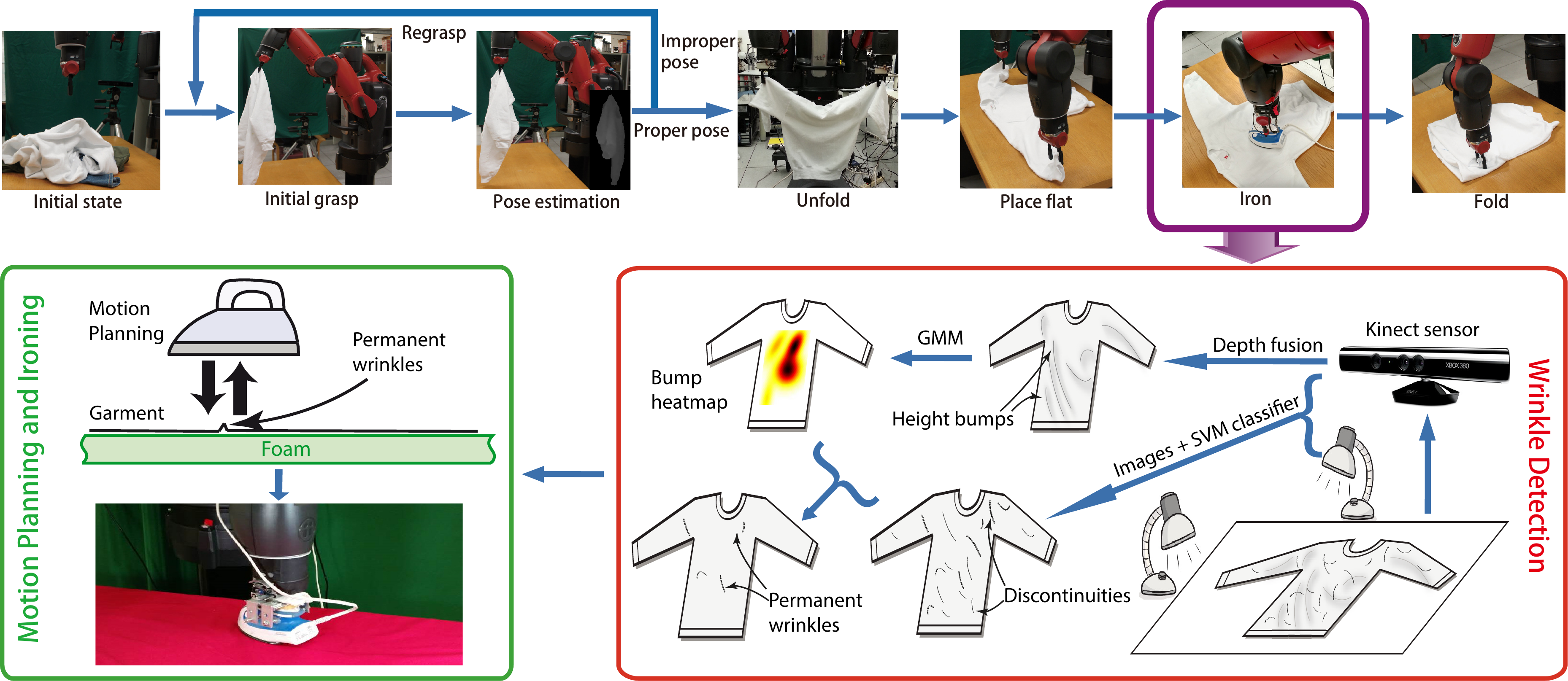}
  \caption{
    {\sc Top Row: } The entire pipeline of dexterous manipulation of deformable objects~\cite{LiICRA2014}\cite{LiIROS2014}\cite{LiICRA2015}\cite{LiIROS2015}.
    In this paper, we are focusing on the phases of garment ironing, as highlighted in the purple rectangle.
    {\sc Bottom Row: } Detailed ironing process from wrinkle detection to motion planing and ironing. We first detect height bumps using the depth fusion algorithm. Then applying GMM to fit those height bumps. We also employ diffuse reflection to detect surface discontinuities. Combining these two results, permanent wrinkles are detected using our probabilistic multi-sensor framework. Finally, all the detected wrinkles are sorted in an optimized way and exported to the Baxter robot for ironing.}
\label{fig:flowchart}
\vspace{-0.3cm}
\end{figure*}

Building a system to do household chores such as laundry is a challenging robotics goal.
One component of this is ironing, which is aimed at removing 
\emph{wrinkles} from fabric, making the garment look more dressy or
formal. 
Figure~\ref{fig:iron_intro} shows our proposed robotic ironing setup.
A Baxter robot is ironing a wrinkle on a piece of cloth, which is detected by a surface curvature scan using a Kinect depth sensor and a discontinuity scan using a Kinect RGB camera and additional light sources.
Figure~\ref{fig:flowchart} shows a complete pipeline of post-laundry garment 
manipulations, starting from grasping, continued by visual recognition, 
regrasping, unfolding, placing flat, ironing, and folding. 
Our previous work~\cite{LiICRA2014}\cite{LiIROS2014}\cite{LiICRA2015}\cite{LiIROS2015} 
has successfully addressed all the stages of the pipeline with the exception of the ironing task. 
This paper adds an ironing component to our existing pipeline.

A garment is typically manufactured by stitching together pieces of fabric,
each of which is cut out from an originally flat fabric sheet. 
As long as the contacts between the weft and warp threads do not slide over each other,
a regular garment is thus locally \emph{developable}, meaning that the Gaussian
curvature (the product of the two principal curvatures) is zero everywhere except along the seams. 
When experiencing forces due to squeezing or swirling motions, the garment may undergo a series of deformations,
resulting in (possibly a mixture of) the following two types of developable
regions: (1) non-smooth regions; (2) smooth regions.

Non-smooth regions usually cannot be flattened by iterative pulling.
We define those regions as \emph{permanent wrinkles}, which are the ones we want to use the iron on.
For the smooth regions, most of them can be flattened by pulling the boundaries iteratively.
We try to remove the permanent wrinkles (purple rectangle in Figure~\ref{fig:flowchart}), through a series of techniques 
described in this paper using an iron mounted on one of the Baxter robot hands.

For a permanent wrinkle smaller than the iron base with its surrounding garment region being 
more or less flat, we found that we can reduce the permanent wrinkle by 
static placement of the iron on the surface or dynamic sliding of the iron onto the surface for larger ones. 
For a region with mixed deformations (i.e., smooth and non-smooth regions), we expect that we can pull the fabric to 
locally remove the developable deformation (by using e.g., \cite{sun2015}).

Permanent wrinkles correspond to high curvatures or discontinuities in the gradient of the garment shape. 
Other deformed regions are much smoother, and their height deviations are larger.
An off-the-shelf sensor, like Kinect, is useful for capturing regions with large variation in height. 
But it is usually too low resolution for capturing smaller surface discontinuities.
On the other hand, because of the rapid change in the gradient (or normal) 
around the permanent wrinkles, there is a rapid change in the lighting effect. 
Therefore, an illumination based approach suffices to capture the permanent wrinkles. 
We found that using a combination of a 3D depth scan and an illumination based approach allows us to robustly identify permanent wrinkles from other deformations without a carefully calibrated system.

We start by roughly placing the garment flat on the table, which can be achieved by ours and others previous 
work~\cite{Towner2011}\cite{DoumanoglouICRA2014}\cite{LiICRA2015}. 
We used a Kinect with a custom two axes mount, which is placed about $80$ cm above the table. 
The axes of the motors are aligned in orthogonal directions to obtain depth images with enough parallax for reconstruction and fusion of multiple scans~\cite{Newcombe2011}. 
We also place two light sources at two adjacent sides of the table, as shown in Figure~\ref{fig:iron_intro}. 
The garment lit by the two light sources will have differences in the lighting effect around the non-smooth regions (permanent wrinkles). 
We combine a 3D height map image and a 2D illumination image in a probabilistic framework, and classify and rank the deformations for ironing. 
The robot aligns the principal axis of the iron with the target wrinkle, computes the trajectory and approach to the garment and performs ironing. 
After several iterations, we obtain a desired ironed result with most of the permanent wrinkles removed. The contributions of our paper are: 
\begin{itemize}
\item[-] A taxonomy of a few common wrinkles on a piece of cloth and garments, as well as the physical-level explanation.
\item[-] A multi-sensor probabilistic framework for classifying detected wrinkles for robotic ironing, which combines curvature-based and discontinuity-based images.
\item[-] Experiments with a Baxter robot showing effective robotic ironing using both static and dynamic ironing.
\end{itemize}

\section{Related Work}
\label{sec:relatedwork}

There are many challenges associated with the manipulation of a deformable object such as a garment.
Many researchers started with recognizing the category and pose of a deformable object using a large database, 
which contains exemplars either from off-line simulation or real garments~\cite{DoumanoglouECCV2014}\cite{LiICRA2014}\cite{LiIROS2014}\cite{Willimon2013}.
By iterative regrasping of the garment by hands by a robot, the garment finally reaches a stable state that can be placed flat on a table~\cite{Towner2011}\cite{LiICRA2015}\cite{StriaIROS2014}.
These methods proceed to garment folding by first parsing of its shape~\cite{LiICRA2015}\cite{Milleretal_IJRR2012}\cite{millerICRA2011}\cite{StriaIROS2014}. 
With the shape parameters, a folding plan can be generated and executed either by a humanoid robot~\cite{LiIROS2015}\cite{vandenberg2010}, or by two industrial arms~\cite{StriaIROS2014}.

What is left in our pipeline (see Figure~\ref{fig:flowchart}, top row) is to have an efficient ironing process, and identifying wrinkles is the first step of it. 
To compute the surface curvature, many researchers proposed methods to extract features from depth visual cues~\cite{Shepard2010}\cite{RamisaICRA2012}\cite{Willimon2013}.
Those detected wrinkles are useful for robotic grasping, regrasping, and object classification.
One notable work that uses detected wrinkles to achieve a flattened surface is by Sun \etal~\cite{sun2015}.
They apply two SLR cameras as a stereo pair to reconstruct a high quality depth map of garment surface.
Then by estimating the volume of the ridges, the robot arms are able to flatten the garment by iterative dragging.

Robotic ironing of deformable garments is a difficult task primarily because of the complex surface analysis, regrasping, and hybrid force/position control of the iron.
Without wrinkle detection, Dai \etal introduced an ironing plan that spreads out the whole garment surface by 
dividing it into several functional regions~\cite{Dai2004_traj}.
For each region, in terms of the size and shape, an ironing plan is automatically generated.
Dai \etal also addressed the ironing problem considering the folding lines~\cite{Dai2004_folding}.

\begin{figure}[!htpb]
\begin{center}
\includegraphics[width=0.48\textwidth]{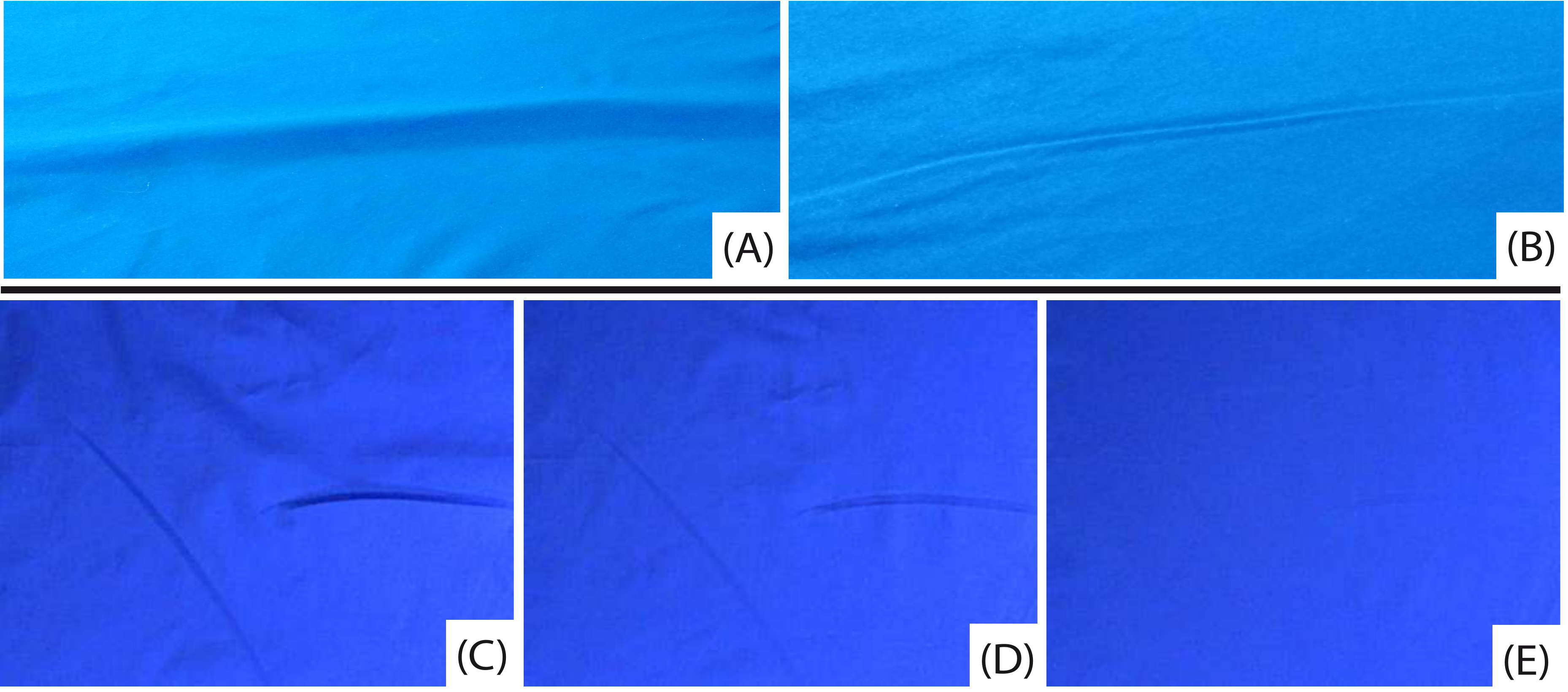}
\vspace{-0.5cm}
\end{center}
   \caption{Best viewed by zooming.
	{	{\sc{top:}} Effects of ironing on a bump. (A): A typical height bump on a piece of cloth. (B): Results of ironing on the height bump.
	\sc{bottom:}} A comparison of effects of pulling the boundaries and ironing on wrinkles. (C): A region with height bumps and permanent wrinkles. (D): After pulling the boundaries of the region, there are still some permanent wrinkles left. (E): The results of our ironing method showing the removal of the permanent wrinkles.
	}
\label{fig:failure_case}
\vspace{-0.3cm}
\end{figure}

Figure~\ref{fig:failure_case} shows some examples of robotic ironing.
The top row shows a case of ironing onto a height bump.
If ironing a height bump, we can remove the air trapped in the bump, but permanent wrinkles could also be created. 
This implies that we need to avoid ironing a height bump. 
The bottom row shows a comparison of pulling the boundaries of the fabric, and ironing for wrinkle removal.
We can see that pulling can only remove some height bumps (smooth regions) but not permanent wrinkles.
But with further ironing, all the permanent wrinkles can be removed.

\section{Problem Formulation}
\label{problem_formulation}
In our approach, we employ two different types of visual scans to detect non-smooth and smooth regions.
One is a curvature scan by a Kinect sensor which fuses multiple depth images to find \emph{height bumps} (see Sec.~\ref{sec:curvature_scan}). 
Height bumps are non-flat regions with smooth curvature.
The other one is a discontinuity scan using diffuse reflection to find \emph{permanent wrinkles} (see Sec.~\ref{sec:wrinkle_detection_diffuse}).
In terms of the scan resolution, the Kinect has relatively lower resolution in the surface reconstruction, whereas the diffuse reflection is able to capture smaller surface discontinuities.
Empirically, the non-smooth regions are mostly the permanent wrinkles as we defined in Sec.~\ref{sec:intro}.
The height deviation of smooth regions detected by the curvature scan are likely height bumps, caused by frictional force between the garment and the table, which can be removed by iterative pulling~\cite{sun2015}.

With the two scan methods mentioned above, we build a probabilistic multi-sensor framework that can efficiently detect and classify the permanent wrinkles that need to be ironed.
Since it is expensive to set up a well-calibrated diffuse reflection system to accurately reconstruct and measure the volume of the wrinkles, we argue that our framework is more generalizable and easy to deploy.

For the discontinuity scan, we are only interested in the position of the permanent wrinkles but not full 3D reconstruction.
The regions to be evaluated here are permanent wrinkles (non-smooth regions), which are from the discontinuity scan. 
Let $\mathcal{K}$ represent a set of the detected height bumps from the curvature scan and $\mathcal{D}$ represent a set of the illumination discontinuities from discontinuity scan.
We first define a pixel-wised probability.
The probability of an illumination discontinuity ${d_i} \in \mathcal{D}$ in a permanent wrinkle set $\mathcal{W}$ is evaluated as,
\begin{equation}
\label{eq:pixel_wise}
\begin{aligned}[rl]
P(d_i \in \mathcal{W}) &= \sum_{d_i} P(x \in \mathcal{W}) \\
&=\sum_{d_i} P(x \in \mathcal{D}) P(x \in {\overline {\mathcal{K}}})R(d_i)
\end{aligned}
\end{equation}
where $x$ represents a pixel in the image, and $R({d_i})$ evaluates the confidence of a discontinuity ${d_i}$ to be classified as a permanent wrinkle in terms of the local shape features, etc.  
Then we further define,
\begin{equation}
\label{eq:qd}
Q({d_i} \cap \mathcal{\overline K}) = \sum_{d_i} P(x \in \mathcal{D}) P(x \in \overline{\mathcal{K}})
\end{equation}
where $Q({d_i} \cap \mathcal{\overline K})$ evaluates the probability of a discontinuity ${d_i}$ in $\mathcal{D}$ not intersecting with height bumps using the results $\mathcal{K}$ from the curvature scan, and $\mathcal{\overline K}$ represents regions other than detected height bumps.
From Eq.~(\ref{eq:qd}), we can infer that $Q({d_i})=1$.
And Eq.~\ref{eq:pixel_wise} can be written as,
\begin{equation}
\label{eq:eq3}
P({d_i} \in {\mathcal{W}}) = Q({d_i} \cap \mathcal{\overline K}) \cdot R({d_i})
\end{equation}

\subsection{Joint probability evaluation}
We could calculate $P(d_i \in \mathcal{W})$ using Eqs.~(\ref{eq:qd}) and~(\ref{eq:eq3}).
For robust ironing, we want to avoid ironing regions that are close to a height bump or in between adjacent height bumps. 
We found that using a Gaussian Mixture Model (GMM) as follows works well to approximate $P(d_i \in W)$ for our purpose.
We start by first using the results from the curvature scan and then the discontinuity scan in the detection pipeline.
The probability of a discontinuity $d_i$ to be a permanent wrinkle given detected height bumps from curvature scan in set $\mathcal{K}$ is evaluated by $Q({d_i} \cap \mathcal{\overline K})$.
Using the Bayes rule, it can be written as,
\begin{equation}
\label{eq:bayes}
Q({d_i} \cap \mathcal{\overline K}) = Q({d_i}|\mathcal{\overline K}) \cdot Q({\mathcal{\overline K}})
\end{equation}
$Q(\mathcal{\overline K})$ is a joint distribution of ${k_1},{k_2}...{k_N} \in \mathcal{K}$. 
We further simplify the scenario that all the height bumps are created independently. Applying the chain rule with conditional dependency assumption,
\begin{equation}
\begin{aligned}[rl]
Q({d_i} \cap \mathcal{\overline K}) &=Q({d_i}|{\overline{k}_1},{\overline{k}_2} \cdot  \cdot  \cdot {\overline{k}_N})Q({\overline{k}_1},{\overline{k}_2} \cdot  \cdot  \cdot {\overline{k}_N}) \\
&= \prod\limits_{j = 1 \cdot  \cdot  \cdot N} {Q({d_i}|{\overline{k}_j})}=\prod\limits_{j = 1 \cdot  \cdot  \cdot N} ({Q({d_i})-Q({d_i}|{k_j})}) \\
&= \prod\limits_{j = 1 \cdot  \cdot  \cdot N} ({1-Q({d_i}|{k_j})})
\label{eq:chain_prob}
\end{aligned}
\end{equation}
Considering the height bumps detected from the curvature scan are mostly smooth regions, we can formulate those regions as an hypothesis for the permanent wrinkles.
The detected height bumps can be represented as several Gaussian distributions.
Eq.~(\ref{eq:chain_prob}) can be written as,
\begin{equation}
\prod\limits_{j = 1 \cdot  \cdot  \cdot N} (1-{Q({d_i}|{k_j})}) = \prod\limits_{j = 1 \cdot  \cdot  \cdot N} (1-{Q({d_i}|\mathcal{N}(k_j^{(x,y)},\Sigma_j^{(x,y)}))}
\end{equation}
where $k_j^{(x,y)}$ is the center of the height bump $k_j$.
The covariance matrix $\Sigma_j^{(x,y)}$ is defined as,
\begin{equation}
\Sigma_j^{(x,y)} = \left[ {\begin{array}{*{20}{c}}
   {{D_1}} & 0  \\
   0 & {{D_2}}  \\
\end{array}} \right]
\end{equation}
where $D_1$ and $D_2$ is the fitted first and the second principal axes of the segmented height bump.
The closer the $d_i$ to the height bump, the higher probability $Q(d_i|k_j)$ gets.
For each detected discontinuity $d_i$, we will evaluate its probability to be ironed as in Eq.~(\ref{eq:chain_prob}).
This formulation can be further approximated by a Gaussian Mixture Model (GMM).

\subsection{Individual probability evaluation}
In this part, we show the inference of $R({d_i})$.
The images captured from the discontinuity scan themselves carry useful information for permanent wrinkle classification, such as texture and edges.
Therefore, we propose a SIFT-SVM based classifier to find permanent wrinkles at pixel level.

The training data is collected by capturing dozens of images with normalization under a Lambertian reflection model (see Sec.~\ref{sec:wrinkle_detection_diffuse} for more details).
Then we manually label those wrinkles which are considered as the permanent wrinkles in the image at pixel level to form a positive training set ${\mathcal{T}_p}$.
We also create a negative training set $\mathcal{T}_n$ by scattering a certain amount of pixel points over the whole image.
Since the wrinkles only occupy a small set of pixels in the image, all the scattered points are mostly negative samples.

For each pixel in training set ${\mathcal{T}_p}$ or ${\mathcal{T}_n}$, we compute the SIFT descriptor~\cite{vedaldi08vlfeat}. Then we create two training sets $\{\bf{X}\}^{|{\mathcal{T}_p}|}$ and $\{\bf{X}\}^{|{\mathcal{T}_n}|}$, where ${\{{\bf{X}}\}^{|N|}}=\{{\bf{x}}_1, {\bf{x}}_2,...{\bf{x}}_N\}$ contains $N$ sets of $128$ dimensional feature vector.
Each pixel will be considered as a training sample in training a SVM classifier.
The output is a classification confidence score that will be assigned to each pixel $I_{u,v}$ as, 
\begin{equation}
S(I_{u,v}) = sigmoid (\sum\limits_{j = 1}^N {{a_j}{y_j}\kappa({x_j},x) + b} )
\end{equation}
where $(a_1, a_2,...,a_N, b)$ are parameters from SVM training, 
and $\kappa({x_j},x)$ is a kernel function.
Then the individual probability of a discontinuity to be a permanent wrinkle is calculated as a normalized summation of pixels that overlaid on itself as,
\begin{equation}
R({d_i}) = \frac{1}{{\left| {{d_i}} \right|}}\sum\limits_{{I_{u,v}} \in {d_i}} {S({I_{u,v}})}
\end{equation}
where $\left| {{d_i}} \right|$ represents number of pixels associated with this discontinuity.

\section{Multi-sensor Detection}
From Sec.~\ref{problem_formulation}, we assume that the garment or a piece of cloth has been roughly flattened on a table, 
with some non-smooth and smooth regions existing.
In this section, we focus on the detection of height bumps and discontinuities over the surfaces.

\subsection{Curvature scan}
\label{sec:curvature_scan}
\subsubsection{Surface Reconstruction}
To obtain a distribution of the height bumps, a Kinect fusion algorithm~\cite{Newcombe2011} is employed.
The Kinect sensor projects infrared light onto the surface of the cloth and receives the reflection by a monochrome CMOS sensor, which is able to capture 3D information of the projected surface.
Considering the resolution of the Kinect depth fusion stream ($640 \times 480$), such fusion process will lose details of some small local curvature.
However, it can robustly capture the fused height deviation over the surface, which is an important hypothesis for the probabilistic multi-sensor wrinkle detection framework, as described in Sec.~\ref{problem_formulation}. 

\begin{figure}[!htpb]
\begin{center}
\includegraphics[width=0.98\linewidth]{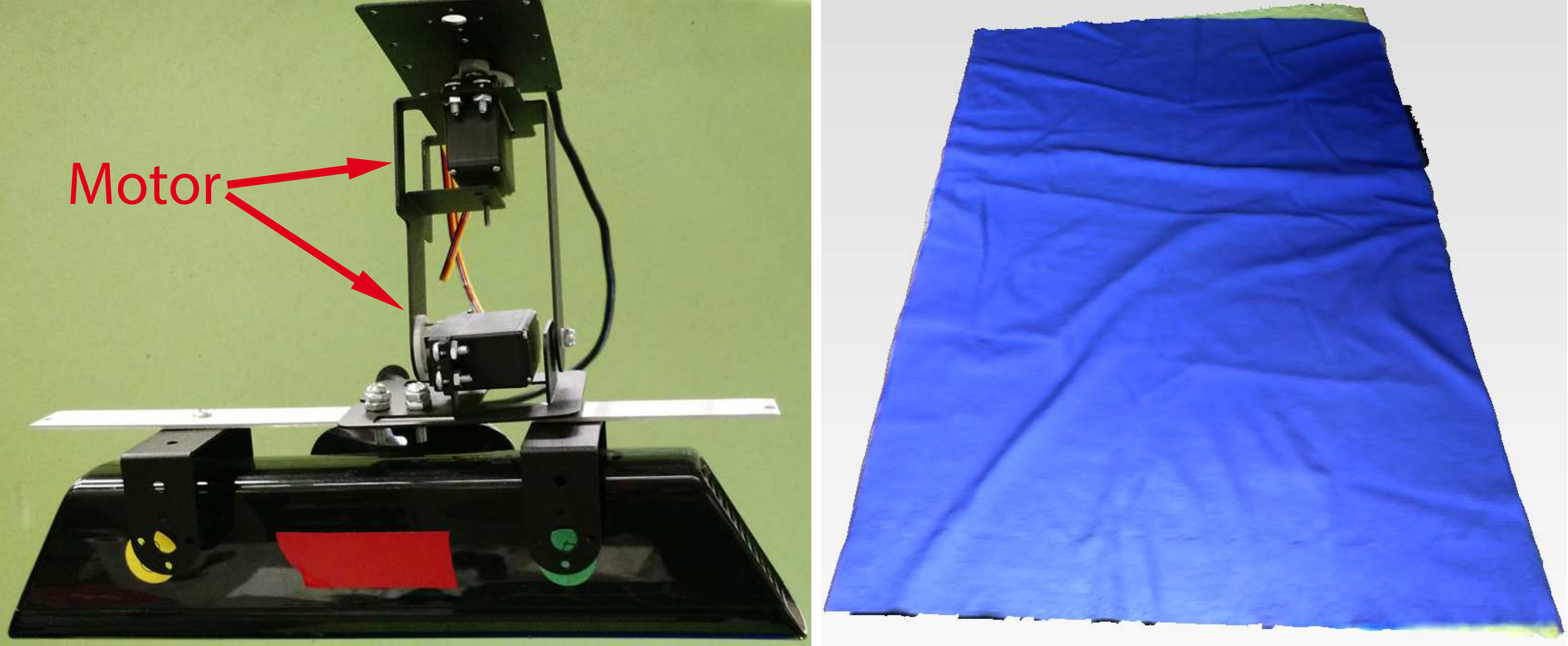}
\end{center}
   \caption{{\sc{left:}} A Kinect attached to two-axis motors rotation module.
	{\sc{Right:}} A reconstructed cloth surface rendered with color. 
	Most of the height bumps are caught, but not the detailed discontinuity information.}
\label{fig:kinect_recon}
\end{figure}

Since the reconstruction requires disparity between each camera frame, simply rotating the Kinect along one axis is not enough.
This will generate little or no parallax between depth frames, which leads to a singularity for a reconstructed 3D point.
We designed a two-axis motor structure to overcome such cases, see Figure~\ref{fig:kinect_recon}.
Each time the two joints will rotate simultaneously and create enough parallax.
Figure~\ref{fig:kinect_recon} right shows a reconstruction result from the setting. 

\subsubsection{Height bump detection}
\label{sec:height_bump}
As seen from Figure~\ref{fig:kinect_recon}, right, the reconstructed mesh has some ridges and rugged terrains.
To find smooth bump-like regions, the \emph{Hessian matrix} is calculated over the whole image $I$ pixel by pixel. 
The second order Hessian matrix $\bf{H}$ is defined as,
\begin{equation}
\bf{H} = \left[ {\begin{array}{*{20}{c}}
   {\frac{{{\partial ^2}I}}{{\partial {x^2}}}} & {\frac{{{\partial ^2}I}}{{\partial x\partial y}}}  \\
   {\frac{{{\partial ^2}I}}{{\partial x\partial y}}} & {\frac{{{\partial ^2}I}}{{\partial {y^2}}}}  \\
\end{array}} \right]
\end{equation}
For each pixel, we can calculate two eigenvalues $\lambda_1$ and $\lambda_2$ of $\bf{H}$.
Assuming $\lambda_1 > \lambda_2$, then pixels that satisfied the following equation will be classified as a bump point and added to a set $\mathcal{R}$~\cite{KOENDERINK1992}.
\begin{equation}
\frac{2}{\pi }{\tan ^{ - 1}}\left( {\frac{{{\lambda _1} + {\lambda _2}}}{{{\lambda _1} - {\lambda _2}}}} \right) \in [ - \frac{1}{8},\frac{5}{8})
\end{equation}
Those connected bump points in $\mathcal{R}$ can be viewed as a height bump. 
Since each pixel is associated with a depth value, the total volume of each region can be estimated by summation of all normalized depth values within the region.
Furthermore, all the bumps can be ranked in terms of their volume, and large volume smooth regions will be defined as the height bumps.
Given a cluster of all ridge points associated with a height bump, the principal axis can be calculated by the PCA algorithm.
Sample results of reconstructed mesh and the detected height bumps will be shown in Sec.~\ref{section:experiments}.

\subsection{Discontinuity scan}
\label{sec:wrinkle_detection_diffuse}
To catch discontinuities over the surface of the cloth, the resolution of the Kinect sensor is not enough.
Therefore, we employ diffuse reflection~\cite{zhang1999shape}.
In Eq. (\ref{eq:bayes}), we have modeled the scenario that the detected height bumps from the curvature scan are considered as the hypothesis for the discontinuity scan.
To generate an efficient ironing plan, we only care about the discontinuity on the surface, which means we do not have to go through a calibration stage.

When using multiple light sources for the reconstruction, calibration of lighting condition is very important.
In terms of the distance from the light source, the decay of illumination may result in missing of a wrinkle detection.
Most of our experimental cloth or garments are made from cotton or natural fabrics, which have the property that the surface's luminance is isotropic. 
For generality, we employ a Lambertian reflectance model~\cite{Oren1995} to formulate the surface luminance.
The Lambertian model can be written as,
\begin{equation}
I(u,v)=a(u,v) \cdot {\bm{n}}(u,v) \cdot {\bm{s}(u,v)}
\end{equation}
where for a pixel location $(u,v)$, $I$ is the intensity of the diffusely reflected light, $a$ is the albedo, $\bm{n}$ is the surface normal, and $\bm{s}$ is the intensity of incoming lights.
For two different light sources, $a$ will be the same, and $\bm{n}$, $\bm{s}$ are different. 
To catch the discontinuity in all directions, two orthogonal light resources are used.

We assume the environmental illumination is negligible when the indoor lights are turned off.
From the camera capture, the current view can be captured as $I_{ref1}$ and $I_{ref2}$ using the first and the second light sources.
With a new piece of cloth, we turn the first and the second light source on, independently, and record the new camera views as $I_1$ and $I_2$.
Given the previous camera observation, a reflection look-up table can be established.
Then the calibrated value of ${{I'}_1}$ and ${{I'}_2}$ can be calculated as,
\begin{equation}
{I'_{1}}(u,v) = {{{I_{1}}(u,v)} \mathord{\left/
 {\vphantom {{{I_{ref}}(u,v)} {{I_{ref1}}(u,v}}} \right.
 \kern-\nulldelimiterspace} {{I_{ref1}}(u,v}}) 
\end{equation}
\vspace{-0.5cm}
\begin{equation}
{I'_{2}}(u,v) = {{{I_{2}}(u,v)} \mathord{\left/
 {\vphantom {{{I_{ref}}(u,v)} {{I_{ref2}}(u,v}}} \right.
 \kern-\nulldelimiterspace} {{I_{ref2}}(u,v}}) 
\end{equation}

The final output image is an combination of images from two light sources by $I = \sqrt {I{'}_1^{2} + I{'}_2^{2}}$.


With the normalized surface image, we run the previous trained SIFT-SVM classifier on each pixel and classify them into permanent wrinkle pixels by the confidence score.
Given all wrinkle pixels, in terms of their connectivity, the Hough transform is applied to find line segments as the detected permanent wrinkles. 
Non-Maximum Suppression is used for avoiding multiple wrinkles at one location.
Sample results are shown in Sec.~\ref{section:experiments}.
\vspace{-0.05cm}

\section{Ironing Procedure}

\subsection{Position-based control vs. force-based control}
Force-based control usually requires a precise force sensor to generate real-time feed back, which is another cost.
Empirically, the ironing task does not require very accurate force feedback to control the path.
Therefore, for the ironing task, we start with an alternative method --- position control.
We place a foam (about 6 cm thick) under the ironing target, with which position-based control can gain similar press effects of force-based control.
Essentially, the foam will passively response to the depth of the iron press and provide a spring-like force feedback; the force will be a linear function of the depth in a simple spring model.
In addition, the position-based control is easier and more convenient for deploying such ironing functionality on a household robot, whereas the force-based control may require a more sophisticated system.  

\subsection{Ironing path planning}

\subsubsection{Ironing orientation}
Most irons are equipped with a V shape head which can easily push the air out from under the wrinkle and flatten the surface.
In our scenario, for each targeted wrinkle, the principal axis is calculated.
Then the end effector of an arm will align the head of the iron with the axis, given that the robot has previously stored the orientation of the iron to its base.

\subsubsection{Ironing motion}
Given the targeted permanent wrinkle, the start and end ironing position can be generated in terms of the center of the iron base.
As mentioned before, we place a foam under the ironing target, so that when the iron is placed lower than the surface of the target, pressure will be created.

We also define two ironing motions, which are static ironing and dynamic ironing.
The length of the wrinkle is transformed back to the world coordinate system, and one of the motions is selected based upon the length.
More specifically, if the length of the wrinkle is less than $70\%$ of the longer axis of the iron, the static ironing motion is selected, otherwise, the sliding motion.
However, as the iron slides on the cloth, it may cause an effect of air accumulation, which may create more permanent wrinkles and height bumps.
Thus, we limit the length of the detected wrinkle in the Hough transform at most twice the length of a iron, and break down a longer one into shorter ones.

\subsubsection{Ironing path optimization}
During the final step of surface analysis, a set of permanent wrinkles $\mathcal{W}$ will be detected, all of which will be ironed.
In our scenario, each permanent wrinkle is represented by a line segment.
For each of the line segment, the iron will move from one end point to the other end point as an iron action.
If a line segment is selected randomly at each time, the whole process could take longer time because the iron may move back and forth.
Therefore, we design an algorithm that minimize such time using a greedy search.
The first selected permanent wrinkle is the one with the highest probability calculated from our framework.
Each time, the iron looks for the closest end point of another line segment for the next ironing process.
				%

\section{Experimental Results}
\label{section:experiments}

To evaluate our results, we tested our method on several different garments and pieces of cloth for multiple trials.
A high resolution video of our experimental results is online at {\url{http://www.cs.columbia.edu/\textasciitilde yli/ICRA2016}}.

\subsection{Robot Setup}
In our experiments, we use a Baxter research robot, which is equipped with two arms with seven degrees of freedom. 
We mount a Kinect sensor~\cite{Zhang_kinect_2012} on top of the table, facing down, which has been calibrated to the robot base frame, as shown in Figure~\ref{fig:kinect_recon}, left.
As described in Sec.~\ref{sec:curvature_scan}, a two-axis module is attached to the Kinect to create enough parallax for depth fusion.

\begin{figure*}[h]
\begin{center}
\includegraphics[width=0.98\linewidth]{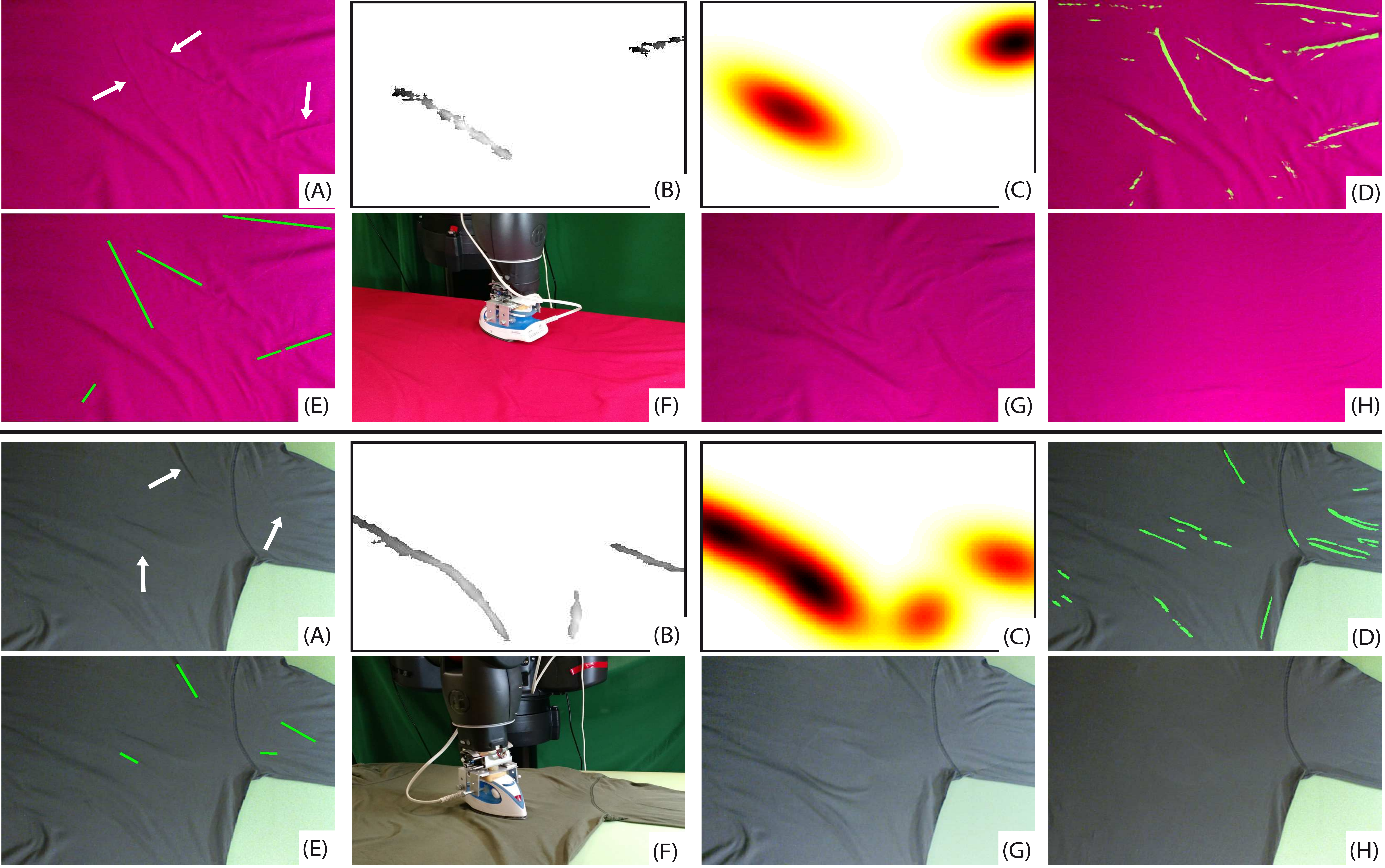}
\end{center}
   \caption{Best viewed by zooming.
	Two examples of whole process of the robotic ironing.
	(A): original roughly flattened cloth (white arrows show some obvious wrinkles). (B): Detected height bumps. (C): GMM fitting results. (D): SVM-based discontinuity detection results. (E): Final permanent wrinkle detection results, which combines result from C and D. (F): The Baxter robot ironing. (G): The ironing result. (H): The ironing result after manually flattening.}
\label{fig:iron_results}
\end{figure*}

\subsection{Curvature scan and discontinuity scan}
The curvature scan aims at detecting large height bumps over the cloth surface.
For each scan, we start the two motors simultaneously for two completed rotations.
The quality of the reconstructed mesh is gradually improved as more scans are taken. 
As described in sec. ~\ref{sec:curvature_scan}, the height bumps are detected by computing the Hessian matrix and volume estimation.
In terms of the bump volume, we discard small ones with a given threshold.

The discontinuity scan aims at detecting shadow discontinuities over the cloth surface.
We manually create a training dataset which contains thousands of pixels on the discontinuity region by manually labeling.
Then we apply the trained SVM classifier on new images.

\subsection{Robotic ironing}
We have tested our approach on several different pieces of cloth and garments.
Here we demonstrate two examples of all the steps in robotic ironing in Figure~\ref{fig:iron_results}.
It starts with the curvature scan to detect height bumps using the Kinect sensor.
The detected bumps are shown in (B). 
Then We apply GMM approximation on (B) given the number of bumps as shown in (C).  
With discontinuity scan (D) and our probabilistic multi-sensor model, permanent wrinkles are detected as shown in (E) in green line segments.
Those wrinkles, represented as line segments, are sorted in a way that robot can iron them with minimum movement and alignment.
(G) shows an image of the cloth after ironing, and (H) is after manually flattening (note that permanent wrinkles cannot be removed by such flattening).
We can see that some discontinuities on or close to the height bumps can be removed in the final permanent wrinkle detection.

\subsection{Ironing with optimized path}
We have tested our ironing path plan on several different garments with more than one permanent wrinkle detected.
When multiple permanent wrinkles are detected, an optimized ironing path is generated, and the iron will traverse all the wrinkles.
An optimized path will reduce the unnecessary alignment and movement of the iron, as well as inverse-kinematics computation.

\begin{figure}[!htpb]
\begin{center}
\includegraphics[width=0.99\linewidth]{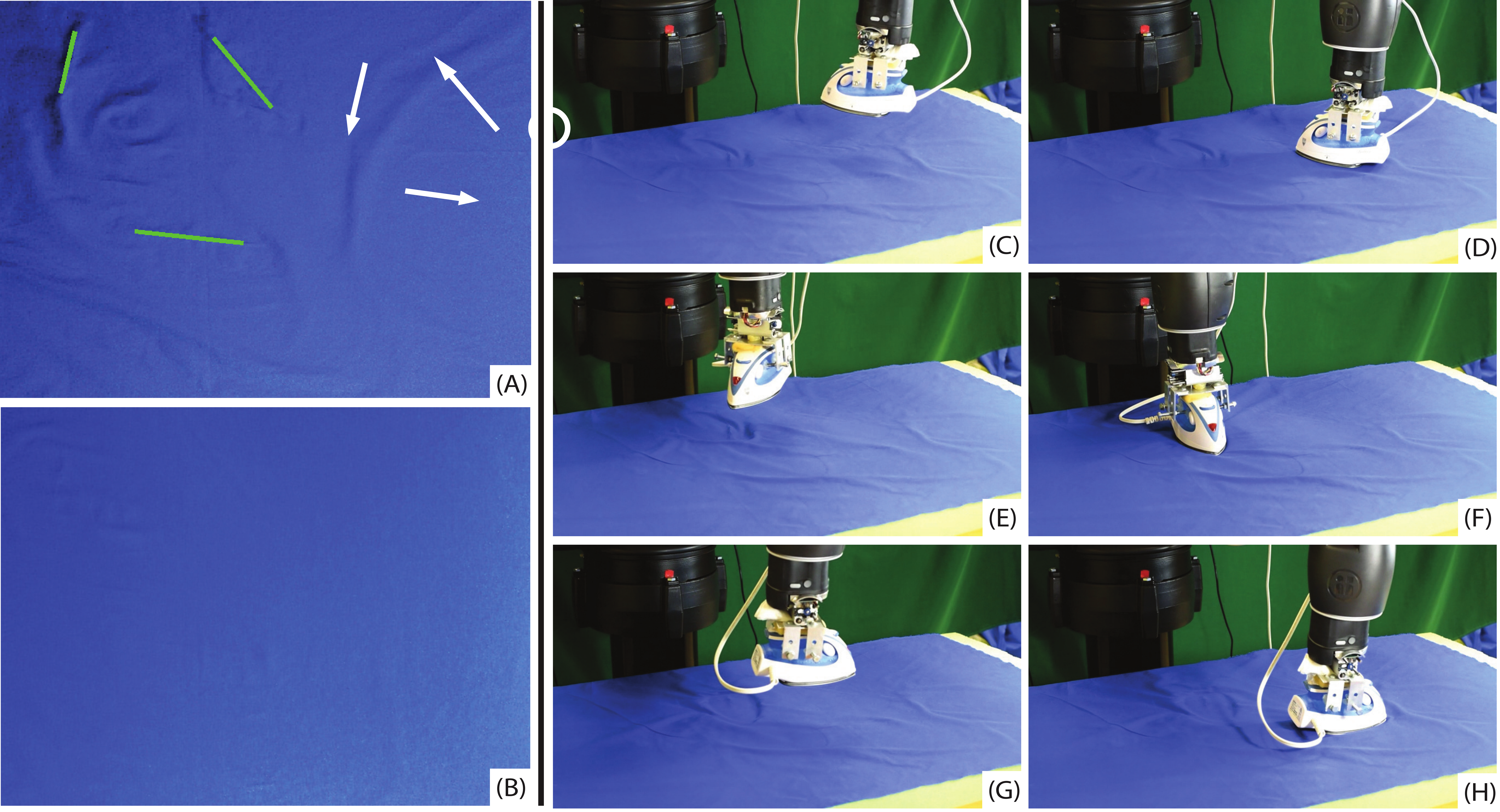}
\vspace{-0.5cm}
\end{center}
   \caption{ An example of robotic ironing with an optimized path. In this case, three permanent wrinkles are detected. 
	(A): Green line segments are detected permanent wrinkles. White circle is the starting position and white arrows show the path orientation.
(B): Results after ironing and manually flattening.
(C)-(H): Key frames of the whole ironing process in an order.
	}
\label{fig:iron_path}
\vspace{-0.1cm}
\end{figure}

Figure~\ref{fig:iron_path} shows an example of the ironing with an optimized path.
(A): A piece of cloth contains several height bumps and permanent wrinkles. Green line segments are detected permanent wrinkles. White circle is the starting position and white arrows show the path orientation.
(B): Results after ironing and manually flattening.
(C)-(H): Key frames of the whole ironing process in an order.
The iron first aligns with the first wrinkle and moves towards the starting position. 
After each ironing, the iron will be lifted up about $5 cm$, and move to the next permanent wrinkle.
In this case, the whole ironing process takes about $35$ seconds, which is faster than a path without optimization or a randomized path.

\section{Conclusion}

In this paper, we have presented a probabilistic multi-sensor framework for wrinkle detection in a robotic ironing task.
We analyze the surface by first using a curvature scan by a Kinect sensor to find height bumps, which should not be included in the ironing. 
Then by using a discontinuity scan with two light sources, and a SVM based classifier to find discontinuities on the surface. 
Combining results from the two scans, permanent wrinkles can be detected and represented as line segments.
A Baxter robot, holding an iron in one hand, irons the surface guided by those detected line segments in an optimized order.
Experimental results show that with our detected wrinkles, the Baxter robot is able to iron and remove them iteratively and produce a wrinkle-free area on the cloth.

Robotic ironing is a very challenging task.  
A full solution to the problem requires complex surface analysis, regrasping, and hybrid force/position control of the iron.   
In this paper, we have addressed the surface analysis problem and also used a position controlled robotic arm to implement ironing.  
While position control can remove many wrinkles, we also believe that with more accurate force control, the effects of ironing can be improved.  
One of our future directions is to add the force control component to our system.
The Lambertian assumption used in the discontinuity scans works for many garments but it is not clear it will work for all garments.

\textbf{Acknowledgments} 
We'd like to thank R. Ying for many discussions. 
We'd also like to thank NVidia Corporation, and Intel Corporation for the hardware support. 
This material is based upon work supported by the National Science Foundation under Grant No. 1217904.


\bibliographystyle{plain}
\bibliography{refs_yinxiao}
\end{document}